\documentclass[11pt,a4paper]{article}
\usepackage[hyperref]{acl2020}
\usepackage{times}
\usepackage{latexsym}
\usepackage{graphicx}

\usepackage{microtype}

\aclfinalcopy 

\title{EmotionGIF-IITP-AINLPML: Ensemble-based Automated Deep Neural System for predicting category(ies) of a GIF response }

\author{
Soumitra Ghosh$^*$, Arkaprava Roy$^+$, Asif Ekbal$^*$ \and Pushpak Bhattacharyya$^*$\\
$^*$Department of Computer Science and Engineering\\
Indian Institute of Technology Patna, Patna, India \\ 
\{1821cs05,asif,pb\}@iitp.ac.in\\
$^+$Department of Electronics and Communication Engineering\\
National Institute of Technology, Durgapur, India \\
apr.17u10296@btech.nitdgp.ac.in\\
}

\date{}

\begin{document}
\maketitle
\begin{abstract}
In this paper, we describe the systems submitted by our IITP-AINLPML team in the shared task of SocialNLP 2020, EmotionGIF 2020, on predicting the category(ies) of a GIF response for a given unlabelled tweet. 
For the round 1 phase of the task, we propose an attention-based Bi-directional GRU network trained on both the tweet (text) and their replies (text wherever available) and the given category(ies) for its GIF response. In the round 2 phase, we build several deep neural-based classifiers for the task and report the final predictions through a majority voting based ensemble technique. Our proposed models attains the best Mean Recall (MR) scores of 52.92\% and 53.80\% in round 1 and round 2, respectively.

\end{abstract}

\section{Introduction}
Identifying emotion in dialogues is one of the most challenging tasks in the area of natural language processing, which is very important for building the dialogue systems \cite{chen2018emotionlines}. Though sentiment analysis is being studied in natural language community for a long time \cite{zhang2018deep}, understanding multiple emotions expressed in chats and conversations comes up with relatively harder challenges \cite{balahur2015sentiment} for many reasons. 

Various individuals may respond differently towards the same comment. Absence of voice modulations and facial expressions in informal chatting makes it difficult to capture emotion in the conversation. This type of problem is earlier addressed in EmoContext \cite{chatterjee2019semeval} and EmotionX-2018 \cite{hsu2018socialnlp}. People during chatting often take help of emojis, figures and message contractions not only to reduce effort and shorten comments but also to express their emotions better. In this case, GIFs play an essential role \cite{miltner2017never}. Often social media users reply with GIFs only, without any text response. Therefore, to understand the different emotions associated with a GIF in the reply, it is necessary to consider the comment and its relation with its associated reply. This type of problem is earlier addressed in EmotionX-2019 challenge \cite{shmueli2019socialnlp} where the task was to predict emotions in spoken dialogues and chats. 

The enhanced and better expressiveness of GIFs in comparison to other popular graphics-based media, such as emojis and emoticons have made their utilization amazingly mainstream on social media and a significant expansion to online human communication which has motivated the introduction of EmotionGIF 2020 \cite{shmueli2020socialnlp} shared task. 

\subsection{Problem Description}

Given an unlabelled tweet and its reply (optional), the challenge is to recommend the possible categories its GIF response may belong to. All tweets in the training set have their GIF responses with some tweets having their text responses as well. The task requires to return a non-empty subset of 1-6 categories from among 43 possible GIF categories for a given unlabelled tweet. 

\begin{table*}
\centering
\begin{tabular}{lllll}
\hline
\textbf{idx} & \textbf{text} & \textbf{reply} & \textbf{categories} & \textbf{mp4} \\
\hline
47 & what is sleep & - & ["shrug", "idk"] & 4f69c84cb....mp4 \\
42 & Are your ready for America & - & ["slow\_clap", "applause", & \\ & to Open Up Again? & &  "yes"] &  9acf47aab3....mp4 \\
95 & we as a collective have been & & & \\ & traumatized by men with J & & & \\ & names. & Woahhhhh lol & ["yes"] & 12d85340f....mp4 \\
\hline
\end{tabular}
\caption{\label{datasamples}
Training data samples. The filenames under 'mp4' field are shortened to accommodate in the limited space. Empty cells under 'reply' field for tweet id 47 and 42 imply that those tweets only have a GIF response.}
\end{table*}

In this paper, we develop a deep learning framework for predicting the categories of a GIF response for an unlabelled tweet. We build multiple deep neural-based systems, such as CNN \citep{kim2014convolutional}, Bidirectional Gated Recurrent Unit (Bi-GRU) \cite{cho2014learning}, and Bidirectional Long Short Term Memory Networks (BiLSTM) \cite{graves2005framewise}. We support our models with an attention mechanism \cite{bahdanau2014neural} that emphasizes on the important parts of a given input tweet. We combine multiple basic models to result in a couple of stacked architectures (Stacked BiGRU, Stacking BiGRU with BiLSTM), and finally, we report the final predictions from a majority voting-based ensemble \cite{ghosh2020cease} method that combines all the developed models. Our proposed frameworks are less complex than the standard transformer models, but can provide reasonably good results and can be trained using local GPU support conveniently.

The rest of the paper is organized as follows: Section 2 gives a brief description of the dataset and the various preprocessing measures applied to them. The details of the proposed methodologies are discussed in section 3. In Section 4, we discuss the various experimental details and their results. Finally, we conclude the paper in section 5.

\section{Datasets}
\subsection{Description}
EmotionGIF 2020 \cite{shmueli2020socialnlp} provides 40,000 two-turn Twitter threads: the original tweet, and the response tweet (which includes an animated GIF), where train set consists of 32,000 and each of development and evaluation set consists of 4,000 samples. Besides, training data is accompanied by the animated GIF responses (for sake of completeness of the data) corresponding to each tweet in the video (.mp4) format. 

Along with the tweets (\textit{Text}) and their text responses (\textit{Reply}), the train instances includes a \textit{Category} field that consists (not for all instances)  single or multiple category labels to which the GIF response belongs. Table \ref{datasamples} shows some sample instances from the training dataset. The text of the original tweet includes mentions (@user), hashtags (\#example), emojis, etc. As tweets containing links have been filtered out, the dataset is left with text-only tweets. The \textit{Reply} column, in general, contains the text of the response and in case of GIF only response, the reply is an empty string. Some samples from the training dataset are shown in Table 1.
\subsection{Pre-processing}
We perform basic cleaning and pre-processing operations (removal of hashtags, URLs and mentions, extra blank spaces removal, conversion to lower case, etc.) on the tweets before using them for our experiments. Smileys are replaced with their meaning like :-) is replaced with the word 'happy'. We use the emoji\footnote{\url{https://pypi.org/project/emoji/}} python library to replace emojis by their meanings. We also build a dictionary of popular contractions and their elongated representation and use them to represent the words in the tweets in their entirety.

\section{Methodology}
In this paper, we present two frameworks employed in two different rounds. We first experiment with the text provided by in the tweets and then include the replies in the experiment to achieve better results. In both the rounds double input models are used for capturing emotions inherent in the tweets and their replies. In first round we employ a deep learning model consisting of bidirectional GRU and attention layers. In second round ensemble learning strategy is followed for improving the performance.

\begin{figure*}[!h]
\begin{center}
\includegraphics[scale=0.48]{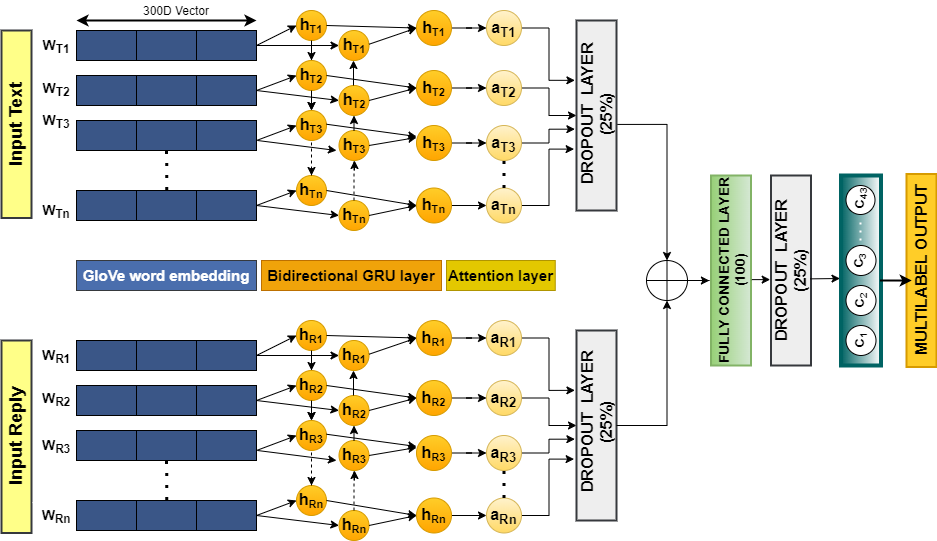} 
\caption{Model architecture for Round 1 submission}
\label{fig:1}
\end{center}
\end{figure*}

\subsection{System description for Round 1}

The overall architecture of our model for the round 1 submission is shown in Figure \ref{fig:1}. We use the pre-trained GloVe\footnote{GloVe: \url{http://nlp.stanford.edu/data/wordvecs/glove.840B.300d.zip}} \cite{pennington2014glove} word embeddings of 300 dimension to initialize the word embedding layer. We extract the text and reply features from two separate but identical branches comprising of the input layer, word embedding layer, Bi-GRU encoding layer, attention layer and a dropout layer. The output from the embedding layer is passed through a BiGRU-based word encoding layer (128 units) that learns the contextual representation of the input followed by an attention layer \cite{bahdanau2014neural} that emphasizes on the important parts of the input. We apply dropout \cite{srivastava2014dropout} of 25\% to the individual attention layer outputs to prevent overfitting. The individual text and reply features for a training row instance is concatenated and passed through a fully connected dense layer (100 neurons) with ReLu activation followed by another dropout layer (25\%). The output layer is a dense layer of 43 neurons (each neuron represent each category) with Sigmoid activation. We use binary cross-entropy as the loss function and employ Adam \cite{kingma2014adam} optimizer to train the network. 

\begin{figure}[!h]
\begin{center}
\includegraphics[scale=0.40]{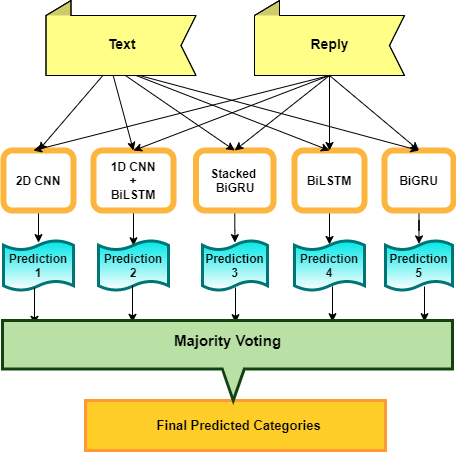} 
\caption{General architecture of the majority voting-based ensemble}
\label{fig:2}
\end{center}
\end{figure}

\subsection{System description for Round 2}
 

We build several deep neural architectures for the task in hand and report the final predictions via a majority voting-based ensemble approach. All the developed architectures for round 2 can be realized easily from the figure \ref{fig:1} itself as the only change has been in the word encoding part.

In our 2D CNN model, the input is passed through four 2D convolutional layers via a 1D spatial dropout layer (40\% dropout). All the convolutional layers have 32 filters with ReLu activation function and kernel weights are generated with the normal distribution. We consider kernel sizes of 1,2,3 and 5 for the different convolutional layers respectively. Each of these convolutional layers is followed by a two-dimensional maxpooling layers. Outputs of the maxpooling layers are concatenated along with the columns and flattened.
Rest of the architecture follows similarly as described in figure \ref{fig:1} from the fully connected layer onwards. 

Our second model is a stacked version of Convolutional and LSTM networks. It has three convolutional (1D) layers which take input from the embedding layer. Each of them has 100 filters with kernel size 2, 3 and 4 and zero padding such that input and output have the same size. Outputs from these layers are added and passed through a 1D maxpooling layer with pool size 2. The output is fed to a bidirectional LSTM layer (128 units). Rest of the architecture follows similarly as described in figure \ref{fig:1} from the input-specific attention layer onwards.

In the next model, we stack two consecutive bidirectional GRU layers (128 units) which takes input from the embedding layer. Rest of the architecture follows similarly as described in figure \ref{fig:1} from the input-specific attention layer onwards.

We replace the GRU cells with LSTM units keeping all other components in figure \ref{fig:1} as it is. This results in our fourth model which we call BiLSTM. We consider the model developed for round 1 submission as the fifth and the final model for building our ensemble framework.

\textbf{Majority Voted system:} For all the aforesaid models, 6 categories are predicted for each development/test instance. Frequencies of the predicted labels from all the 5 systems are calculated and a sorted list of distinct labels is created in decreasing order of frequency values. The first 6 elements of the sorted list are considered as the final GIF categories for a development/test instance.

\section{Experiments, Results and Discussion}
\subsection{Experimental Setup}
We use the popular python-based libraries Keras\footnote{\url{https://keras.io/}} \cite{chollet2015keras} and Scikit-learn\footnote{\url{https://scikit-learn.org/stable/}} \cite{pedregosa2011scikit} with Tensorflow backend and GPU support for our experiments. We evaluate our models using the Mean Recall at k, with k=6 (MR@6) score which is the official metric of EmotionGIF 2020 \cite{shmueli2020socialnlp}. For internal evaluation of the developed models, we split the training data in 80:20 ratio and train our model on the larger split and evaluate on the smaller one.

\begin{table}[!h]
\centering
\begin{tabular}{llll}
\hline
\textbf{Round} & \textbf{MR\textsubscript{Dev}} & \textbf{MR\textsubscript{Eval}} & \textbf{Rank} \\
\hline
1 & 52.46\% & 52.92\% & 5\\
2 & 53.63\% & 53.80\% & 5\\
\hline
\end{tabular}
\caption{\label{results}
MR scores on development (Dev) set and test (Eval) set of the shared task. MR: Mean Recall}
\end{table}

\subsection{Results and Discussion}
We evaluate our proposed system for round 1 directly on the development set provided in the shared task. We use the Grid Search method to set the various hyperparameters of the network like the number of neurons in the GRU layer and linear layer. The results for round 1 and round 2 are shown in table \ref{results}. We achieve a MR of 52.92\% (Rank 5) on the test set with the top-performing system (Rank 1) attains a score of 62.47\%. 

We perform two-levels of evaluation of our experimented systems in round 2. First, we consider 20\% of the train data as an interim test set and evaluate the individual model's performance. We consider the top-performing systems for building the majority voting-based ensemble framework. Second, we evaluate the performance of the ensemble framework on the test set and report the predictions as the final output. It can be observed from table \ref{results} that our ensemble-based framework attains a MR score of 53.80\% which is 0.88\% more than what we have achieved in round 1. We also narrow the gap between the performance of the top-performing system (Rank 1 attains 61.74\% MR) for round 2 and our proposed approach.

\section{Conclusion}
In this work, we build an ensemble deep learning framework on top of several attention-based deep neural networks to achieve the task objective of predicting categories for a GIF response. We effectively incorporate both the tweets and their text responses in building our automated systems. Our participation in EmotionGIF 2020 has been a wonderful learning experience for our team as we have achieved 5\textsuperscript{th} rank in both the rounds with attained MR@6 scores of 0.5292 and 0.5380, respectively. We look forward to learn more from the best-performing systems. Results indicate that our models can serve as strong baselines as an alternative framework to transformer-based approaches. 

In future, we will try to enrich the learning of our developed end-to-end systems by effectively incorporating multimodal features extracted from the GIFs of training data and map them to unlabelled test data (using text similarity measures on the given tweets and text replies of the train and test sets).

\bibliography{acl2020}
\bibliographystyle{acl_natbib}

\end{document}